# A Novel data Preprocessing method for multi-dimensional and non-uniform data

Farhana Javed Zareen[a,] ∗ and Suraiya Jabin[a]

[a]*Department of Computer Science, Jamia Millia Islamia, New Delhi-110025, India*

## ABSTRACT

We are in the era of data analytics and data science which is on full bloom. There is abundance of all kinds of data for example biometrics based data, satellite images data, chip-seq data, social network data, sensor based data etc. from a variety of sources. This data abundance is the result of the fact that storage cost is getting cheaper day by day, so people as well as almost all business or scientific organizations are storing more and more data. Most of the real data is multi-dimensional, non-uniform, and big in size, such that it requires a unique pre-processing before analyzing it. In order to make data useful for any kind of analysis, pre-processing is a very important step. This paper presents a unique and novel preprocessing method for multi-dimensional and non-uniform data with the aim of making it uniform and reduced in size without losing much of its value. We have chosen biometric signature data to demonstrate the proposed method as it qualifies for the attributes of being multi-dimensional, non-uniform and big in size. Biometric signature data does not only captures the structural characteristics of a signature but also its behavioral characteristics that are captured using a dynamic signature capture device. These features like pen pressure, pen tilt angle, time taken to sign a document when collected in real-time turn out to be of varying dimensions. This feature data set along with the structural data needs to be pre-processed in order to use it to train a machine learning based model for signature verification purposes. We demonstrate the success of the proposed method over other methods using experimental results for biometric signature data but the same can be implemented for any other data with similar properties from a different domain.

## 1. Introduction

Most recently, various industries have started using sensors for their daily operations. When multiple sensors record data, the values are continuous, multi-dimensional and each dimension is of varying length. The data is stored in its raw form for analyzing it later. One such data is Biometric Signature data which is recorded using specialized devices with the help of many sensors acting together while a signer signs. Imagine an analysts trying to design biometric signature verification system by making use of such data with lot of variables or dimensions with varying length. This is very pertinent problem before any data analyst ahead of analyzing the data. In order to acquire the sample biometric signature data, the person is asked to write his/her name a number of times in the signature capturing device during the enrolment phase. After all the samples of an individual are collected, all of its important features are extracted from it for example the x, y co-ordinate of the signature, pen tilt angle, the velocity with which the person signs, the total time taken by the signer, the acceleration, the number and position of pen-ups and pen-downs, number of strokes etc. This is a very significant and important step in biometric signature verification since it is required to find the most significant features of a signature that minimizes intrapersonal variations and maximizes interpersonal variations [7]. Now, the most important issue of a biometric sign; every time a person signs on the digitizer, he/she signs with slight or major variations, which we call intra class variation i.e. the difference in the features of the genuine samples provided by an authentic individual. For example at one point the total time taken by a signer (one biometric feature of sign) is 10 seconds and at some other point of time the total time taken by the same signer may be 12 seconds and accordingly values will be recorded for other biometric features also giving another sign of same user a different dimension. Therefore the number of sample points that we get each time is of different length even for the same signer as we can see from Table 1.

It is difficult to analyze this data with different number of feature search time the same sample is being trained and therefore data pre-processing in this case is very important. In this paper we present a unique method that will pre-process the non-uniform and inconsistent data from the same signer that is captured from the digitizing tablet and convert it into a uniform data set which can be further easily used to train the recognition system.

The complexity of data increases manifold with the increase in dimensionality or number of features. For instance, biometric signature verification data of m signers, each person's sign having n features(x, y co-ordinate of the signature, pen tilt angle, velocity with which the person signs, timestamp, the number and position of pen-ups and pen-downs, number of strokes etc.) and p number of samples are collected per signer. These n features are captured at fixed timestamp but depending upon the time taken by a signer the number of points collected are different for different samples even for same user. For example at an attempt, the user can take 1 second to sign but at a later attempt, the same user might take 1.2 seconds to sign and therefore the number of data points collected may vary and thus the biometric signature data; say 20 signatures, collected for one user is with varying dimensions. So, the data is not only multidimensional but also the different dimensions are non-uniform. Other examples of such data sets may include gene expression data, sequence data, proteomics data and pathway data etc. An analysis is program would take enormous time to process such data and remain failure in answering complex queries, which could not be visualized beforehand. Among the various goals addressed by data pre-processing, dimensionality reduction or feature selection has been recognized as a central problem in data analysis [8].

The most popular and linear data transformation method is principal component analysis (PCA), but many nonlinear dimensionality reduction techniques also exist [9],[10]. For multidimensional data, tensor representation can also be used in dimensionality reduction through multi-linear subspace learning [11].

The principal component analysis technique performs a linear mapping of the data to a lower-dimensional space in such a way that the variance of the data in the low-dimensional representation is maximized. In practice, the original space (with dimension of the number of points) is reduced (with data loss, but retaining the most important variance) to the space spanned by a few eigenvectors.

For high-dimensional datasets (i.e. with number of dimensions more than 10), dimension reduction is usually performed prior to applying a K-nearest neighbor's algorithm(k-NN) in order to avoid the effects of the curse of dimensionality [12].

The present paper aims at presenting a novel method for dimensionality reduction of biometric signature verification dataset such that there is minimal loss of data and the quality of data is also not compromised.

| Table 1. Details of a signature sample | | | | | | | | |
|---|---|---|---|---|---|---|---|---|
| Signature Sample for user 1 | | | | | | | | |
| S.No. | x-coordinate | y-coordinate | Timestamp | Button status-penup/down | Azimuth | Altitude | Pressure | Forged/Genuine |
| 1. | 2288 | 7111 | 75748770 | 0 | 1310 | 680 | 244 | Genuine |
| 2. | 2252 | 7058 | 75748780 | 1 | 1320 | 650 | 247 | |
| . | . | . | . | . | . | . | . | |
| . | . | | | | | | | |
| 500 | . | | | | | | | |
| Signature Sample for user 1 but done by some imposter | | | | | | | | |
| S.No. | x-coordinate | y-coordinate | Timestamp | Button status | Azimuth | Altitude | Pressure | Forged/Genuine |
| 1. | 2288 | 7111 | 75748770 | 0 | 1310 | 680 | 244 | Forged |
| 2. | 2252 | 7058 | 75748780 | 1 | 1320 | 650 | 247 | |
| . | . | . | . | . | . | . | . | |
| . | . | | | | | | | |
| 270 | . | | | | | | | |

## 2. Data Acquisition

### 2.1. Data capturing devices

Today's technology supports a broad range of signature capturing devices that not only captures the shape/picture of the signature but also records the other online/dynamic features of a signature such as pen pressure on the pad, velocity with which a person is signing, tilt angle of the pen, number of strokes etc. It records the spatial as well as temporal information of a signature[13]. This biometric smart pen (BiSp) is a pen that allows recording the hand movements and gestures made by fingers on a signature pad or sometimes even in air [14]. This device is equipped with a number of sensors that can sense the grip of fingers and even a small tilt of angle while holding the pen. However different signature capturing devices acquire different features depending on the specification of the device. Some of the devices are listed below:

- Pen Pads: It consists of a pen and a pad, and is especially designed for capturing signature along with its static and biometric characteristics. Some of its variations are:
  - Wacom Pen Tablet used in [15], [16] and [17].
  - BiSp used in [14]
  - XGT Serial Digitizing tablet used in [18].

- Tablet PC: Tablet PCs are also being used for capturing online features of signatures. Two different models of Tablet Pc have been used in [19].
  - Hewlett-Packard TC 1100
  - Toshiba Portege M200

Both of these devices acquire the discrete time series of x, y coordinates of the signature and the pressure.

- Smart Phones: [20] have used iPhone to capture the dynamic information of the signatures. There are many signature capturing devices that function once attached to a smart phone for example grabba, it works for iPhone and Samsung galaxy (source: www.grabba.com/). Accelerometer and gyroscope are two hardware that are present in almost all the smart phones. Accelerometer in mobile devices is used to find out the orientation of the phone. Gyroscope estimates the angular rotational velocity. These devices can be used to measure the pen-tilt angles, the movement of the hands, the direction of the tablet etc.[21]

- Data Glove: Data Glove is an input device which is worn in hand and it captures all the information of the hand gesture of an individual during the signing process. It can record the following characteristics

Distinct patterns depending on the signers hand size and signature.
  - Total time taken in the process of signing.
  - Hand trajectory dependent rolling.
  - 5DT Data Glove 14 Ultra has been used in [22].

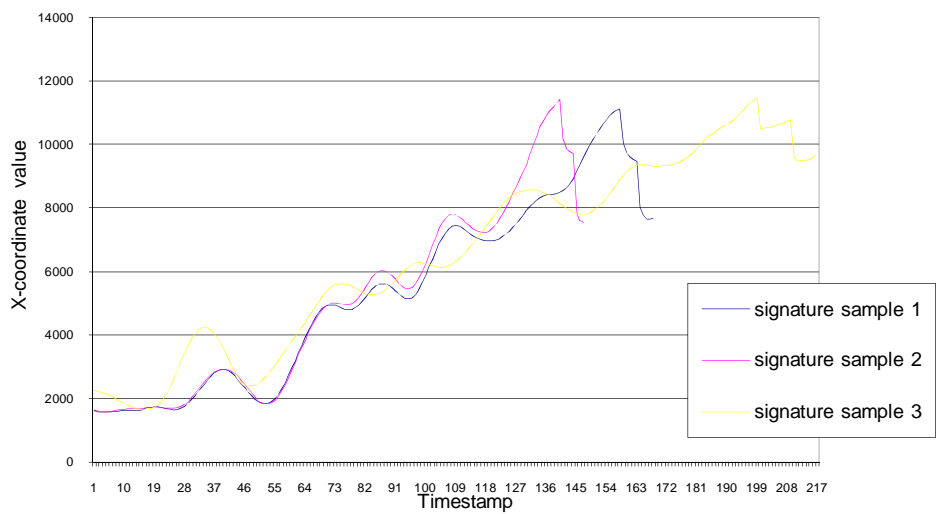

**Fig. 1.** Plot of the x-co-ordinate values of three signature samples of the same signer.

*2.2. Data Collection*

We have used the standard dataset from SVC 2004, available at https://www.aut.bme.hu/Pages/Research/Signature/Resources. It consists of 20 original and 20 forged signatures for 200 signers. Each sample from each signer contains variable data points which are a result of the intra class variations of a signature. The data points in each sample ranges from 200 to 300 and are of varying dimensions. So there is an emergent need to efficiently pre-process the data in order to make it uniform as well as preserving all the biometric features. This paper presents a unique and effective mathematical solution to this problem.

*2.3. Feature Extraction*

- The database contains seven dimensional features which are
- X-coordinate - the co-ordinate of x-axis at each time stamp.
- Y-coordinate - the co-ordinate of y-axis at each time stamp.
- Timestamp - system time at fixed interval of time
- Button status - 0 for pen-up and 1 for pen-down
- Azimuth - clockwise rotation of cursor about the z-axis
- Altitude - angle upward toward the positive z-axis
- Pressure - adjusted state of the normal pressure

At fixed interval of time the value of all of the above features are captured for each signer's each signature sample. The duration of the signing process of different individual may be different, even it may be slightly different for same signer's different signatures. In effect, every time a signer signs, there remains a slight variation in the duration of the signing process. This creates a sort of inconsistency in data which the proposed paper attempts to resolve. Fig. 1 shows the difference of the length the data samples of the same individual. In this figure, the first feature (one out of the seven features) is plotted for three samples of one signer from the database. We can see from the plot that all the three data sets are of different length. The x-axis represents the timestamp. Timestamp is the fixed interval of time at which the data is recorded, and the y-axis represents the value of the x-coordinate at that time interval. From the plot we can see that all the datasets (even if they belong to the same signer) have different sizes which make it difficult to train the dataset.

In the present paper, a novel method is proposed for pre-processing such data set with varying dimensions and each with varying length.

**3. The proposed method for pre-processing of non-uniform multidimensional data curve fitting technique**

In the proposed work, we suggest a unique mathematical modeling of datasets possessing properties as described in the previous section by fitting the data on an appropriate curve such that some specific and fixed no. of points can be extracted as essential features of the data set, and thus making the data set size uniform and homogeneous.

*3.1. Non-linear curve fitting*

A non-linear equation can be represented as equation 1:

$$Y = f(X, \alpha) + \varepsilon \tag{1}$$

Where Y represents the data that is to be modeled, $X = (x_1, x_2, ..., x_n)'$ is the independent variable, $\alpha = (\alpha_1, \alpha_2, ..., \alpha_n)'$ is the model parameter and $\varepsilon$ is the errors or residuals.

The goal of this non-linear curve fitting step is to find an estimation of the parameter values that most appropriately describes the data points. The idea is to minimize the errors between the experimental data and the theoretical data. This is done using chi-square ($\chi^2$) minimization method as given by equation 2. This method is a reasonable method to approximate the data points of a curve that fits a dataset. The square of the difference between the predicted values and the actual values is summed up and is divided by the variance of the data. If the value of Chi-square is near to 1 then we get a reasonably good fit and if it is much greater than 1 then the fit is poor[23].

$$\chi^2 = \sum_N \{\frac{1}{\sigma_i^2}[y_i - f(x_i, \alpha)]^2\} \quad 0 < \chi^2 < \infty \tag{2}$$

Here, N is the number of data points, $\sigma_i^2$ is the variance related to the measurement error for $y_i$, $y_i$ is the observed mean and $f(x_i)$ is the predicted mean [8].

To estimate $\alpha$ using the least square method, we need to solve the normal equation 3

$$\frac{\delta \chi^2}{\delta \alpha} = -2 \sum_{i=1}^{N} \frac{1}{\sigma^2}[y_i - f(x_i, \alpha)][\frac{\delta f(x_i, \alpha)}{\delta \alpha}] = 0 \tag{3}$$

Since explicit solutions cannot be derived from this therefore an iterative method has been applied to find the value of the parameters with an aim to reduce the $\chi^2$. When the $\chi^2$ value in two successive iterations become minimum to a pre-defined threshold then the fitting procedure converges [23].

In order to apply $\chi^2$, the first task is to find the type of function that would be appropriate for fitting a curve to the data. Four typical functions that have been most frequently used for curve fitting of all sorts of data in the literature [24],[25],[26] are:

a) **Weibull Distributions:** The probability density function of Weibull distribution can be expressed as equation 4.[27]

$$f(x) = \frac{\gamma}{\alpha}(\frac{x-\mu}{\alpha})^{(\gamma-1)} \exp(-((x-\mu)/\alpha)^\gamma) \qquad x \geq \mu; \gamma, \alpha > 0 \tag{4}$$

Where,
$\gamma$ is the shape parameter,
$\mu$ is the location parameter and
$\alpha$ is the scale parameter.

Case in which $\mu = 0$ is called a 2-parameter Weibull distribution [Lai et al. 2003] which can be expressed in terms of its survival function as given in equation 5.

$$F(t; \beta, \lambda) = \exp[-(\lambda t)^\beta] \tag{5}$$

b) **Sum of Sines Model:** This model fits the periodic functions which is represented as equation 6[28].

$$f(x) = \sum_{i=1}^{n} A_i \sin(B_i x + C_i) \tag{6}$$

Where the parameter A, B, C indicate,
A is the amplitude,
B is the frequency and
C is the phase constant

c) **Polynomial functions:** A polynomial curve with degree k in the Euclidean plane is represented as equation 7[28].

$$P = \{(x,y) \in \Re^2 : y = a_1 x^k + a_2 x^{k-1} + ... + a_k x + a_{k+1}, a_1 \neq 0\},$$

Where $a_1, a_2, ..., a_{k+1} \in \Re$. (7)

A discrete polynomial function can be defined as equation 8,

$$D = \{(x,y) \in Z^2 : 0 \leq y - f(x) \leq w\},$$

Where $f(x) = a_1 x^k + a_2 x^{k-1} + ... + a_k x + a_{k+1}$. \tag{8}

w is a real valued constant. $a_i$ is called the coefficient, k is called the degree and w is called the width of the polynomial curve that has been discretized. w is the distance between the two polynomials y=f(x) and y=f(x)+w [29].

d) **Fourier series:** The Fourier series describes a periodic signal which can be represented in one of the two forms either exponential or trigonometric, it is represented as the sum of sine and cosine functions as described in equation 9[28].

$$y = a_0 + \sum_{i=1}^{n} a_i \cos(iwx) + b_i \sin(iwx) \tag{9}$$

Where $a_0$ is the intercept, w is the fundamental frequency of the signal and n is the number of terms.

e) **Parabolic curve:** A parabola is a curve where any point is at an equal distance from: a fixed point (the focus), and a fixed straight line (the directrix). It can be represented using equation 10.

$y^2 = 4ax$ (10)

We split the feature $X = (x_1, x_2, ..., x_n)'$ into sub-sections that contain 20 data points in order to assess the data closely. For the curve plotted using the first feature of the sample from our database we get the sub-sections shown in figure 2b.

We let the curves that we acquired from our sample be represented as f(x) and the standard curves be represented as g(x), to find the region between these two curves we divide the region into n number of vertical rectangles. Let us take the ith rectangle and find its thickness, say $\Delta x_i$ is the thickness of the rectangle. Let us take a value from the mantissa say xi. If we subtract the lower function from the upper function we get the area of the rectangle from equation 11:

$$A = (f(x_i) - g(x_i))\Delta x_i \tag{11}$$

And by summing up the areas of all the rectangles, we get an approximated area between these two curves using equation 12.

$$A^{c_k} \approx \sum_{i=1}^{n}(f(x_i) - g(x_i))\Delta x_i \qquad (12)$$

We take the limit over the function as shown in equation 13.

$$A^{c_k} = \lim_{\Delta x_i \to 0}\sum_{i=1}^{n}(f(x_i) - g(x_i))\Delta x_i \qquad (13)$$

$\Delta x_i$ is the $i^{th}$ sub-interval's length $(x_i - x_{i-1})$.

$$A^{c_k} = \int_{a}^{b}(f(x) - g(x))dx \qquad (14)$$

Equation 14 represents the Riemann sum.

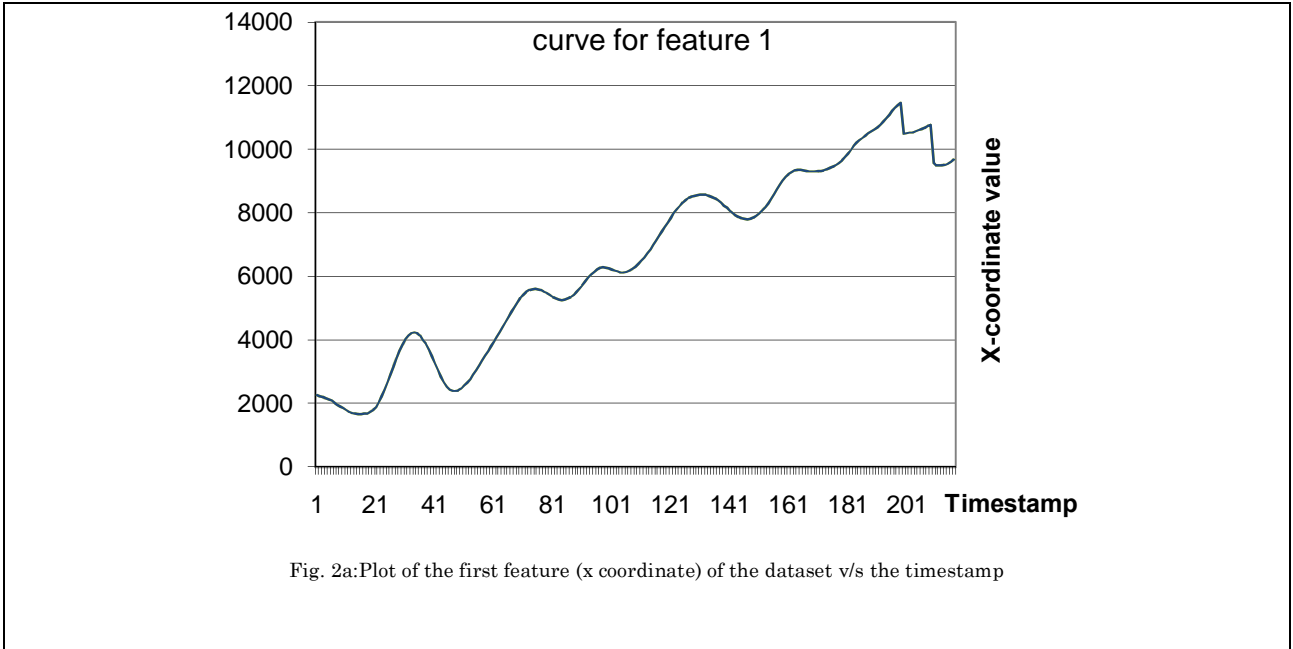

Fig. 2a: Plot of the first feature (x coordinate) of the dataset v/s the timestamp

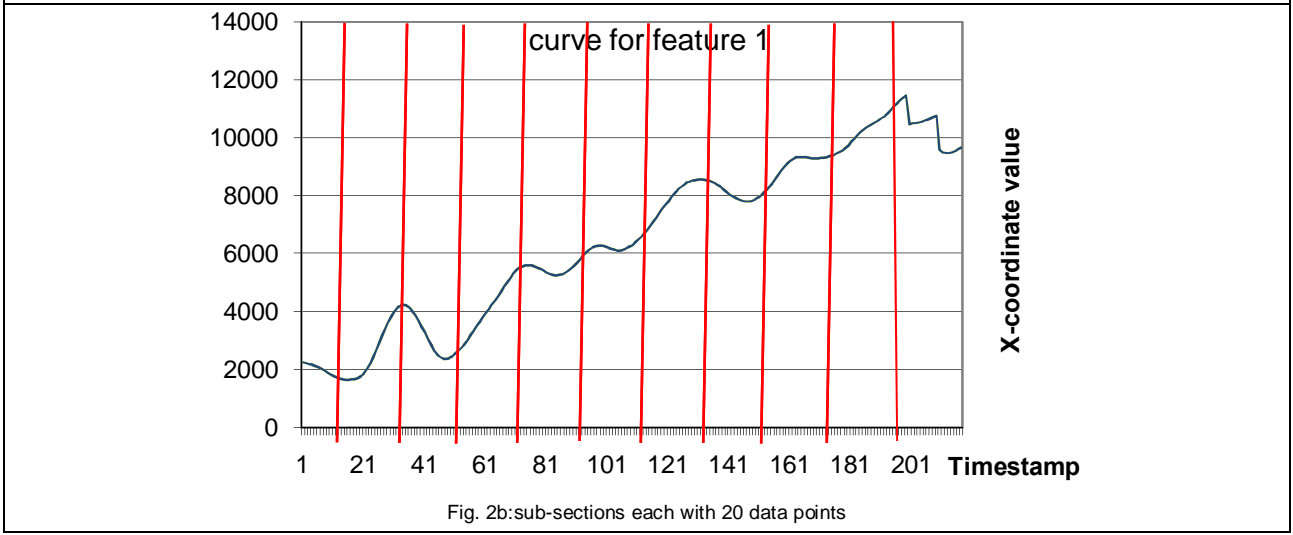

Fig. 2b: sub-sections each with 20 data points

If we take the integration in the interval [0, 1] then we get the equation 15.

$$\int_{0}^{1}1 dx = \int_{0}^{1}(f(x_i) - g(x_i))dx = \lim_{\Delta x_i \to 0}\sum_{i=1}^{n}(f(x_i) - g(x_i))\Delta x_i \qquad (15)$$

Now, we compute the area between the sample curve(plotted from data of a feature) and the standard curves to be fitted using this method for sub-sections plotted for a feature or in general for a dimension of the multi-dimensional data and then we take a summation over this and find out the total area between the curves as shown in fig. 3.

$$A^c = \sum_{j=1}^{k} A^{c^k} \qquad (16)$$

Using equation (16) we compute the areas as shown in table 2:

| Table 2. Computed area between curves | | |
|---|---|---|
| Curves | Equation | Computed area |
| 1. Sinusoidal | $y = \sin x$ | 2.486e+006 |
| 2. Exponential | $y = e^x$ | 5.308e+009 |
| 3. Parabolic | $y^2 = 4ax$ | 3.336e+009 |

We can see that a sinusoidal curve is more efficient as a function to fit the data set (SVC) chosen, into that curve. So, we propose an algorithm using the above results to fit the sample data into a sinusoidal curve. It can be mathematically proved that polynomial function is the best curve fitting function option for timestamp feature as we can plot this feature as a straight line. An algorithm is designed to perform the pre-processing on the data which is given in section 3.2. The same algorithm can be modified to choose any other curve than sinusoidal curve to fit the dataset chosen, following the proposed mathematical data analysis in this section.

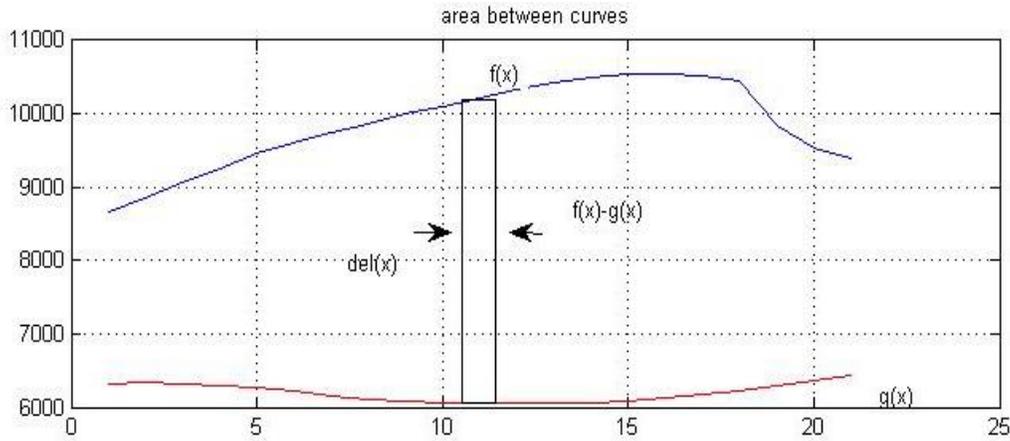

Fig. 3. Area between actual and the fitted curve

*3.2. Proposed Algorithm for Pre-processing of the Data*

**ALGORITHM 1.** Preprocess_SigFeatures

**Input:** Userx_y[m][d] is the input file, it represents $y_{th}$ sample of signature data of User x. It contains m data points of d features same as mentioned in section 2.3.
**Output:** Output[n], array of the sampled data points.
**Begin**
**For** *i=1 to d* **do**
*Userx_y[i]=Userx_y[m][i]*
//extract individual features in 1 D array
**EndFor**
**For** *i=1 to d* **do**   *Output[i]=Curvefitting_SumofSineFunction(Userx_y[i])*
// extract the points after fitting the features in a curve
**If** i=3 **do**
        *Output[i]=Curvefitting_Polynomial(Userx_y[i])*
        //as the third feature (timestamp) performs better with polynomial curve fitting
    **EndIf**
    **Continue**
**EndFor**
**End**

We performed the proposed curve fitting technique for pre-processing the biometric signature data set chosen and obtained the following results as shown in table 2. From the results, we can see that sum of sine functions and polynomial performs better than Fourier and Weibull for pre-processing of biometric signature data.

The Trust Region Algorithms are a set of new algorithms which are used in a number of optimization problems. In this algorithm a region is trusted to approximate the model and this region is called the trust region. The main features of Trust Region algorithm are that it can be used to optimize badly conditioned problems and it has strong convergence properties [30].

The Gauss-Newton method uses the idea of linearizing the non-linear regression function using Taylor series expansion of an unknown vector. Therefore in this method a non-linear problem is converted into a simple quadratic one [4]. Levenberq-Marquardt algorithm is a modification of Gauss-Newton method; it computes the trial step by equation 17.

$$d_k = -(J(x_k)^T J(x_k) + \mu_k I)^{-1} J(x_k)^T F(x_k) \tag{17}$$

The problem arising from singularity and near singularity of $J(x_k)$ is removed by introducing the parameter $\mu_k$ [Jin-Yan 2003].

### 4. Experimental Results and performance evaluation:

For each sample (genuine and forged) of each individual, we perform the four curve fitting technique and we report the errors of these fits on the data used. As we can see from Table 3, Weibull performs a poor fit for the data. But Fourier, Sum of Sine Functions and polynomial give considerable fit for the data. By performing different simulations, we observed that polynomial curve does not perform better than sinusoidal function but for timestamp parameter, polynomial curve outperforms all the other methods that's why in the proposed algorithm the polynomial function is used for timestamp feature and sum of sine functions for the remaining features.

| Table 3. Experimental Results ||||||
|---|---|---|---|---|---|
| Type of Fit | Algorithm | SSE | R-square | Adjusted R-square | RMSE |
| Fourier | Trust Region | 1.333e+007 | 0.9904 | 0.9892 | 315.4 |
|  | Levenberq-Marquardt | 1.332e+007 | 0.9904 | 0.9892 | 315.3 |
|  | Gauss-Newton | 1.171e+007 | 0.9916 | 0.9905 | 295.6 |
| Weibull | Trust Region | 5.308e+009 | -2.816 | -2.841 | 5949 |
|  | Levenberq-Marquardt | 5.308e+009 | -2.816 | -2.841 | 5949 |
|  | Gauss-Newton | 5.308e+009 | -2.816 | -2.841 | 5949 |
| Sum of Sin Functions | Trust Region | 2.486e+006 | 0.9982 | 0.9979 | 139.4 |
|  | Levenberq-Marquardt | 2.486e+006 | 0.9982 | 0.9979 | 139.4 |
|  | Gauss-Newton | 2.488e+006 | 0.9982 | 0.9979 | 139.5 |
| Polynomial | - | 3.854e+007 | 0.9723 | 0.9705 | 521 |

In table 3, we assess the root mean square error (RMSE) for the four methods and from this; we can conclude that Sum of Sine functions outperforms all the other method with minimum RMSE for application of the proposed method on biometric signature data. Trust Region and Levenberq-Marquardt gives similar results which can come out different for different data sets but one can apply the proposed method step by step as described in section 3 for some other data also. Gauss-Newton method gives relatively higher error in comparison with the other two methods.

The results summarized in the table prove that sum of sin functions out performs all the other methods for the dataset chosen (SVC) for experiments and it provides accurate curve for the experimental data, similarly for pre-processing of some other set of data, similar way a suitable curve can be fitted using the proposed method expressed as a generic algorithm of data pre-processing as follows: Suppose there are n data samples with k dimensions each with varying length.

(i)Pick up a dimension of a sample and convert it into fixed length data points say 'p' with curve fitting by selecting an appropriate curve.
(ii)Repeat step (i) for all k-dimensions of the data set.
(iii)The resulting dataset will be of dimension n X p.

After preprocessing the data using sum of sine functions of equation 6, we get an output data in the form of coefficients i.e. a, b and c. We take n to be 11 since we divided our curve into 11 parts. Hence we get 11(the curve is divided into 11 sub curves)*3(three co-efficients A, B, C)*7(seven features of sample) =231. The output and pre-processed, uniform data expressed in terms of coefficients ($A_i$, $B_i$, $C_i$ for i = 1 to 7 features) of the fitted curves for different features, can be represented as shown in table 4.

| Table 4. Pre-processed uniform biometric signature dataset | | | | | |
|---|---|---|---|---|---|
| | $A_1$ | $B_1$ | $C_1$ | ... | $C_n$ |
| User1 | 7.086e+006 | 2.486e+006 | 3.406e+006 | ... | 1.396e+006 |
| User2 | 6.076e+006 | 3.123e+006 | 1.432e+006 | ... | 2.386e+006 |

The result that is obtained through experiments is compared with other pre-processing methods that have been used in the literature for biometric signature data and is summarized in table 5. Here n is the number of samples and d is number of data points per sample.

| Table 5. Comparison with other methods in the literature | | |
|---|---|---|
| Reference | Pre-processing Methods used for biometric signature data | Time complexity |
| C. Boutsidis, M.W. Mahoney, and P. Drineas, 2009 | Singular Value decomposition-feature selection (SVD-FS) | $O(nd \min\{n,d\})$ |
| C. Boutsidis, A. Zouzias, and P. Drineas, 2010 | Random projections-feature extraction | $O(nd\lvert \varepsilon^{-2} k / \log(d) \rvert)$ |
| P. Drineas, A. Frieze, R. Kannan, S. Vempala, and V. Vinay, 1999 | Singular Value decomposition-feature extraction(SVD-FE) | $O(nd \min\{n,d\})$ |
| **Proposed method** | Non-linear curve fitting | $O(d)$ |

## 5. Validation

In order to verify effectiveness of proposed pre-processing method, we performed different simulations by training Bayesian regularized feed-forward neural network using pre-processed biometric signature dataset. Multiple experiments were performed by taking different topologies of the feed forward network along with different parameters like topology, training function and transfer/output function. The best performance obtained along with network features is summarized in table 6 and figure 4.
Figure 4 shows the ROC plot of third simulation from table 7 pictorially using the backpropagation method on the pre-processed biometric signature dataset.

| Table 6. Simulations using Backpropagation Method and Feed-forward neural network | | | | |
|---|---|---|---|---|
| **Training Function** | **Adaption Learning Function** | **Performance Function** | **Topology** | **Transfer Function** |
| trainrp | learngdm | MSE (Mean Squared Error) = 1.2E-03 | 231-60-60-1 | logsig |
| trainbr | learngdm | SSE (Sum Squared Error)=0.004 | 231-70-70-1 | tansig |
| Trainscg | learngdm | SSE (Mean Squared Error) =3.7Ee-09 | 231-50-50-1 | logsig |

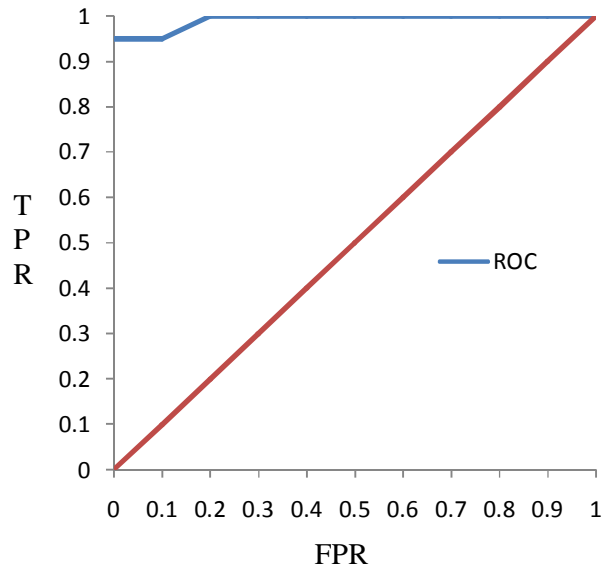

Fig. 4: ROC obtained with the experimental results

Fig. 4 depicts the ROC curve obtained from the experimental results which is a trade-off of true positive rate and the false positive rate.

We conclude that the proposed method can be used as a full proof pre-processing method to reduce the dimensionality of the data set by preserving all biometric features of the data.

And the calculated False Acceptance Rate and False Rejection Rate for the best simulation is:
FAR=0.2625%
FRR=0.5065%

Authors have used various methods to perform biometric signature authentication. The results of these approaches are listed in table 7.

In table 7, we tried to summarize the preprocessing method used by different authors but the authors did not mention anything about how they prepared the data before applying GMM, HMM or DTW etc. methods to design biometric signature based authentication systems. It is worth mentioning that biometric signature data is multidimensional and each dimension is of different length, so it is not suitable to use it as it is with any machine learning technique as mentioned in table 7. The quality of training data certainly affects the designed end system. We interestingly point out that a biometric signature verification system designed with biometric signature data pre-processed using proposed method provides better results as compared to the other methods in terms of EER.

Therefore, it can be deduced that performing preprocessing on biometric signature data enhances the accuracy for authentication in terms of EER, though invariably it's also equally dependent on the machine learning method chosen.

| Table 7. Comparison of different existing authentication systems based onbiometric signatures in terms of EER | | | |
|---|---|---|---|
| Authors | Preprocessing Method used | Machine Learning method/model | Results in terms of EER |
| Lopez-Garcia et al. (2014) | Not mentioned | Gaussian Mixture Model (GMM) | 2.74 |
| Rúa, and Castro (2012) | Not mentioned | Hidden Markov Model (HMM) | 1.83 |
| Bashir and Kempfa (2012) | Not mentioned | Dynamic Time Warping (DTW) | 4.11 |
| **Farhana et al. (2016)** | **Proposed preprocessing method based on non-linear curve-fitting** | **Feed-forward Bayesian regularised Neural Network** | **0.3845** |

## 6. Conclusions

Data pre-processing is often ignored but very important step in any kind of data analytics and the phrase "Garbage In, and Garbage Out" is especially applicable to biometric signature verification data sets. A novel method is proposed to pre-process the inconsistent data that we get from capturing the dataset of biometric signatures. This method can be successfully used to pre-process the varying dimensional data in a consistent and compact form which further can be used in any biometric authentic system for verification. The proposed method's robustness and accuracy is also demonstrated with experimental results. Not necessarily for biometric signature data, this method can be generalized to pre-process any high-dimensional data set.